\documentclass[runningheads,a4paper]{llncs}
\usepackage{amsmath} \usepackage{amstext} \usepackage{amsbsy}
\usepackage{prettyref} \usepackage{latexsym} \usepackage{amssymb}
\usepackage{graphicx} \usepackage{hyperref} \usepackage{braket}
\usepackage{aas_macros}

\let\llncssubparagraph\subparagraph
\let\subparagraph\paragraph
\usepackage[compact]{titlesec} \titlespacing{\section}{0pt}{*3}{*2}
\titlespacing{\subsection}{0pt}{*1}{*0}
\titlespacing{\subsubsection}{0pt}{*0}{*0}
\let\subparagraph\llncssubparagraph

\DeclareMathOperator*{\tr}{tr}

\newrefformat{alg}{Algorithm~\ref{#1}}
\newrefformat{eq}{Equation~\ref{#1}}
\newrefformat{lem}{Lemma~\ref{#1}}
\newrefformat{def}{Definition~\ref{#1}}
\newrefformat{cor}{Corollary~\ref{#1}}
\newrefformat{chp}{Chapter~\ref{#1}}
\newrefformat{sec}{Section~\ref{#1}}
\newrefformat{apx}{Appendix~\ref{#1}}
\newrefformat{tab}{Table~\ref{#1}} \newrefformat{fig}{Fig.~\ref{#1}}

\newcommand{\field}[1]{\mathbb{#1}} 
\newcommand{\Z}{\field{Z}}

\titlerunning{Physical Completeness and Objectivity} \title{Ultimate
  Intelligence Part I: Physical Completeness and Objectivity of
  Induction}

\author{Eray \"Ozkural}

\date{\today}

\begin{document}

\setlength{\abovedisplayskip}{0.11cm}
\setlength{\belowdisplayskip}{0.11cm}

\mainmatter % start of an individual contribution

% \maketitle

\institute{G\"{o}k Us Sibernetik Ar\&Ge Ltd. \c{S}ti.\\
}

\maketitle

\begin{abstract}
  We propose that Solomonoff induction is complete in the physical
  sense via several strong physical arguments. We also argue that
  Solomonoff induction is fully applicable to quantum mechanics.  We
  show how to choose an objective reference machine for universal
  induction by defining a physical message complexity and physical
  message probability, and argue that this choice dissolves some
  well-known objections to universal induction. We also introduce many
  more variants of physical message
  complexity based on energy and action, and discuss the ramifications of our    proposals.
 %\footnote{This paper was initially submitted to
  % ALT-2014. We are taking the valuable opinions of the reviewers
  % into account.}
\end{abstract}

``If you wish to make an apple pie from scratch, you must first invent
the universe.'' -- Carl Sagan

\section{Introduction}

Ray Solomonoff has discovered algorithmic probability and introduced
the universal induction method which is the foundation of AGI theory
% \cite{alp1,alp2,Sol:97discovery}.
\cite{Sol:97discovery}.
% In the present philosophical paper, we investigate the limits of
% intelligence in our physical universe.
Although the theory of Solomonoff induction is somewhat independent of
physics, we interpret it physically and try to refine the
understanding of the theory by thought experiments given constraints
of physical law.  First, we argue that its completeness is compatible
with contemporary physical theory, for which we give arguments from
modern physics that show Solomonoff induction to converge for all
possible physical prediction problems. Second, we define a physical
message complexity measure based on initial machine volume, and argue
that it has the advantage of objectivity and the typical disadvantages
of using low-level reference machines.  However, we show that setting
the reference machine to the universe does have benefits, potentially
eliminating some constants from algorithmic information theory (AIT)
and refuting certain well-known theoretical objections to algorithmic
probability.  We also introduce a physical version of algorithmic
probability based on volume and propose six more variants of physical
message complexity.

\section{Background}

Let us recall Solomonoff's universal distribution.  Let $U$ be a
universal computer which runs programs with a prefix-free encoding
like LISP.  The algorithmic probability that a bit string $x \in
\{0,1\}^+$ is generated by a random program $\pi \in \{0,1\}^+$ of $U$
is:
\begin{equation}
  \label{eq:alp}
  P_U(x) = \sum_{U(\pi) = x(0|1)^*} 2^{-|\pi|}
\end{equation}
% $P_U(x)$ considers any continuation of $x$, taking into account
% non-terminating programs. $P_U$ is also called the universal prior
% for it may be used as the prior in Bayesian inference, for any data
% can be encoded as a bit string.  We shall denote it by merely $P$ in
% the rest of the paper, when we can discern probability of bit
% strings from probability measures.
We also give the basic definition of Algorithmic Information Theory
(AIT), where the algorithmic entropy, or complexity of a bit string $x
\in \{0,1\}^+$ is defined as
% \begin{equation}
%  \label{eq:algo-entropy}
$H_U(x) = \min( \{ |\pi| \ | \ U(\pi)=x \} )$.
%\end{equation}
%We shall now briefly recall the well-known Solomonoff induction
% method \cite{alp1,alp2}.
Universal sequence induction method of Solomonoff works on bit strings
$x$ drawn from a stochastic source $\mu$.  \prettyref{eq:alp} is a
semi-measure, but that is easily overcome as we can normalize it.  We
merely normalize sequence probabilities
% \begin{alignat}{3}
%   \label{eq:normalization}
%   P'_U(x0)=\frac{P_U(x0).P'_U(x)}{P_U(x0)+P_U(x1)} && \quad
%   P'_U(x1)=\frac{P_U(x1).P_U'(x)}{P_U(x0)+P_U(x1)}
% \end{alignat}
eliminating irrelevant programs and ensuring that the probabilities
sum to $1$, from which point on $P'_U(x0|x) = P'_U(x0)/P'_U(x)$ yields
an accurate prediction. The error bound for this method is the best
known for any such induction method.  The total expected squared error
between $P'_U(x)$ and $\mu$ is
% \begin{equation}
%   \label{eq:convergence}
%   E_P \left[ \sum_{m=1}^n{(P'_U({a_{m+1}=1}|a_1a_2...a_m) -
%       \mu({a_{m+1}=1}|a_1a_2...a_m))^2}  \right] 
%   \leq - \frac{1}{2} \ln{P_U(\mu)} 
% \end{equation} which is
less than $-1/2\ln{P'_U}(\mu)$ according to the convergence theorem
proven in \cite{solcomplexity}, and it is roughly $H_U(\mu)\ln2$
\cite{solomonoff-threekinds}.
% Naturally, this method can only work if the algorithmic complexity
% of the stochastic source $H_U(\mu)$ is finite, i.e., the source has
% a computable probability distribution. The convergence theorem is
% quite significant, because it shows that Solomonoff induction has
% the best generalization performance among all prediction methods. In
% particular, the total error is expected to be a constant independent
% of the input, and the error rate will thus rapidly decrease with
% increasing input size.

\section{Physical Completeness of Universal Induction}

Solomonoff induction model is known to be complete and incomputable.
\prettyref{eq:alp} enumerates a non-trivial property of all programs
(the membership of a program's output in a regular language), which
makes it an incomputable function. It is more properly construed as a
semi-computable function that may be approximated arbitrarily well in
the limit. Solomonoff has shown that the incomputability of
algorithmic probability does not inhibit its practical application in
any fundamental way, and emphasized this often misunderstood point in
a number of publications.

The only remaining assumptions for convergence theorem to hold in
general, for any $\mu$ are a) that we have picked a universal
reference machine, and b) that $\mu$ has a computable probability
density function (pdf).  The second assumption warrants our attention
when we consider modern physical theory. We formalize the
computability of $\mu$ as follows:
\begin{equation}
  \label{eq:comp-pdf}
  H_U(\mu) \leq k, \exists k \in \Z 
\end{equation}
which entails that the pdf $\mu(x)$ can be simulated on a computer,
while $x$ are (truly) stochastic. This condition is formalized
likewise in \cite{Hutter:03spupper}.

\subsection{Evidence from physics}

There is an exact correspondence of such a construct in physics, which
is the quantum wave function. The wave function of a finite quantum
system is defined by a finite number of parameters (i.e., complex
vector), although its product with its conjugate is a pdf from which
we sample stochastic observations. Since it is irrational to consider
an infinite quantum system in the finite observable universe, $\mu$
can model the statistical behavior of matter for any quantum
mechanical source. This is the first evidence of true, physical
completeness of Solomonoff induction we will consider. Von Neumann
entropy of a quantum system is described by a density matrix $\rho$:
\begin{equation}
  \label{eq:vonneumann}
  S = -\tr(\rho ln \rho) = -\sum_j \eta_j \ln \eta_j
\end{equation}
where $\tr$ is the trace of a matrix and $\rho=\sum_j\eta_j\ket{j}
\bra{j} $ is decomposed into its eigenvectors. Apparently, von Neumann
entropy is equivalent to classical entropy and suggests a computable
pdf, which is expected since we took $\rho$ to be a finite matrix.
Furthermore, the dynamic time evolution of a wave function is known to
be unitary, which entails that if $\mu$ is a quantum system, it will
remain computable dynamically. Therefore, if $\mu$ is a quantum system
with a finite density matrix, convergence theorem
% \prettyref{eq:convergence}
holds.

The second piece of evidence from physical theory is that of universal
quantum computer, which shows that any local quantum system may be
simulated by a universal quantum computer \cite{Lloyd-Universal}.
Since a universal quantum computer is Turing-equivalent, this means
that any local quantum system may therefore be simulated on a
classical computer.  This fact has been interpreted as a physical
version of Church-Turing thesis by the quantum computing pioneer David
Deutsch, in that 'every finitely realizable physical system can be
perfectly simulated by a universal model computing machine operating
by finite means' \cite{Deutsch-Universal}.  As a quantum computer is
equivalent to a probabilistic computer, whose outputs are
probabilistic after decoherence, these two facts together entail that
the pdf of a local quantum system is always computable. Which yields
our second conclusion.  If $\mu$ is a local quantum system,
% \prettyref{eq:convergence}
the convergence theorem holds.

The third piece of evidence from physics is that of the famous
Bekenstein bound and the holographic principle. Bekenstein bound was
originally conceived for black holes, however, it applies to any
physical system, and states that any finite energy system enclosed
within a finite volume of space will have finite
entropy:\begin{equation}
  \label{eq:bekenstein}
  S \leq \frac{2\pi k R E}{\hbar c}
\end{equation}
where $S$ is entropy, and $R$ is the radius of the sphere that
encloses the system, $E$ is the total energy of the system including
masses, and the rest are familiar physical constants. Such a finite
entropy readily transforms into Shannon entropy, and corresponds to a
computable pdf. The inequality \label{eq:bekenstein} is merely a
physical elucidation of \prettyref{eq:comp-pdf}.  Therefore, if $\mu $
is a finite-size and finite-energy physical system, the convergence
theorem
% \prettyref{eq:convergence}
holds.

Contemporary cosmology also affirms this observation, as the entropy
of the observable universe has been estimated, and is naturally known
to be finite \cite{frampton-entropy-universe}. Therefore, if
contemporary cosmological models are true, any physical system in the
observable universe must have finite entropy, thus validating the
convergence theorem.
% \prettyref{eq:convergence}

Thus, since we have shown wide-reaching evidence for the computability
of pdf of $\mu$ from quantum mechanics, general relativity, and
cosmology, we conclude that contemporary physical science strongly and
directly supports the universal applicability of the convergence
theorem. In other words, it has been physically proven.

\subsection{Randomness, computability and quantum mechanics}

% Recall that the induction operator is an infinite mixture of all
% matching computations, another way to view it is as a mixture of
% pdf's.
Wood et. al interpreted algorithmic probability as a "universal
mixture" \cite{wood-sol}, which is essentially an infinite mixture of
all possible computations that match the input.  This entails that it
should model even random events, due to Chaitin's strong definitions
of algorithmic randomness \cite{chaitin-ait}.  That is to say, the
universal mixture can model white noise \emph{perfectly} (e.g.,
$\mu(x0)=\mu(x1)=1/2$).  More expansive definitions of randomness are
not empirically justifiable.
% since they refer to uncountable sets. No such extra power exists in
% the universe, and such claims must be attributed to a priori
% assumptions such as axiom of infinity.  common among mathematicians
% that do not depend on empirical evidence (e.g., axiom of infinity).
% It is therefore impossible to imagine what more is needed.
Our analysis is that stronger definitions of randomness are not needed
as they would be referring to halting oracles, which would be truly
incomputable, and by our arguments in this paper, have no physical
relevance.  Note that the halting probability is semi-computable.

The computable pdf model is a good abstraction of the observations in
quantum mechanics (QM). In QM, the wave function itself has finite
description (finite entropy), with unitary (deterministic) evolution,
while the observations (measurements) are stochastic.  Solomonoff
induction is complete with respect to QM, as well, even when we assume
the reality of non-determinism -- which many interpretations of QM do
admit.  In other words, such claims that Solomonoff induction is not
complete could only be true if and only if either physical
Church-Turing thesis were false, or if hypercomputers (oracle
machines) were possible -- which seem to be equivalent statements.
The physical constraints on a stochastic source however rules out
hypercomputers, which would have to contain either infinite amount of
algorithmic information (infinite memory), or be infinitely fast, both
of which would require infinite entropy, and infinite energy. A
hypercomputer is often imagined to use a continuous model of
computation which stores information in real-valued variables. By AIT,
a random real has infinite algorithmic entropy, which contradicts with
the Bekenstein bound (\prettyref{eq:bekenstein}).  Such real-valued
variables are ruled out by the uncertainty principle, which places
fundamental limits to the precision of any physical quantity --
measurements beneath the Planck-scale are impossible. Hypercomputers
are also directly ruled out by limits of quantum computation
\cite{Lloyd:Ultimate}. In other words, QM strongly supports the
stochastic computation model of Solomonoff.

\section{On The Existence of an Objective $U$}

% \subsection{Invariance theorem and choice of reference machines}
The universal induction model is seen to be subjective, since the
generalization error depends on the choice of a universal computer $U$
as the convergence theorem shows. This choice is natural according to
a Bayesian interpretation of learning as $U$ may be considered to
encode the subjective knowledge of the observer.  Furthermore,
invariance theorem may be interpreted to imply that the choice of a
reference machine is irrelevant. However, it is still an arbitrary
choice. A previous proposal learns reference machines that have good
programs short in the context of universal reinforcement learning
\cite{Sunehag-agi14}.
% Furthermore, invariance theorem may be interpreted to imply that the
% choice of a reference machine is irrelevant, since any universal
% computer $U_1$ may simulate any other universal computer
% $U_2$. Recall that the invariance theorem is that the algorithmic
% complexity changes at most by a constant when a universal reference
% machine changes.
% \begin{equation}
%   \label{eq:inv-thm}
%   H_{U_1}(x) \leq H_{U_2}(x) + H_{U_1}(s), \iff \forall x, U_2(x) = U_1(sx)
% \end{equation}
% where the simulation program $s$ simulates $U_2$ on $U_1$.

% Nevertheless, in the real world, there is a sense in which LISP is a
% more natural reference machine than C++ in at least two ways:
% \begin{enumerate}
% \item The constants in AIT are smaller for LISP than C++
% \item The programs in C++ are much longer for many ``natural''
%   functions, i.e., LISP captures our intuitive notion of complexity
%   better.
% \end{enumerate}
% And even though LISP seems better for AIT than C++ in practice,
% simulating C++ in LISP requires a long program, which diminishes the
% usefulness of \prettyref{eq:inv-thm}. Therefore, it is reasonable to
% ask in which well-defined sense LISP is better than C++, and whether
% we can define an optimal $U$.

\subsection{The universe as the reference machine}
In the following, we shall examine a sense which we may consider the
best choice for $U$.  Solomonoff himself mentioned such a choice
\cite{solomonoff-theoryandapps}, explaining that he did find an
objective universal device but dismissed it because it did not have
any prior information, since subjectivity is a desirable and necessary
feature of algorithmic probability.
% \begin{quote}
%   For quite some time I felt that the dependence of ALP on the
%   reference machine was a serious flaw in the concept, and I tried
%   to find some ``objective'' universal device, free from the
%   arbitrariness of choosing a particular universal machine. When I
%   thought I finally found a device of this sort, I realized that I
%   really didn't want it - that I had no use for it at all! Let me
%   explain:

%   In doing inductive inference, one begins with two kinds of
%   information: First, the data itself, and second, the a priori data
%   - the information one had before seeing the data. It is possible
%   to do prediction without data, but one cannot do prediction
%   without a priori information. In choosing a reference machine we
%   are given the opportunity to insert into the a priori probability
%   distribution any information about the data that we know before we
%   see it.

%   If the reference machine were somehow “objectively” chosen for all
%   induction problems, we would have no way to make use of our prior
%   information. This lack of an objective prior distribution makes
%   ALP very subjective - as are all Bayesian systems
% \end{quote}
We proposed a philosophical solution to this problem in a previous
article where we made a physical interpretation of algorithmic
complexity, by setting $U$ to the universe itself
\cite{ozkural-compromise}.  This was achieved by adopting a physical
definition of complexity, wherein program length was interpreted as
physical length.  The correspondence between spatial extension and
program length directly follows from the proper physicalist account of
information, for every bit extends in space. Which naturally gives
rise to the definition of physical message complexity as the volume of
the smallest machine that can compute a message, eliminating the
requirement of a reference machine.  There are a few difficulties with
such a definition of complexity whose analysis is in order.  Contrast
also with thermodynamic entropy and Bennett's work on physical
complexity \cite{zurek1998,bennett-physical}.

\subsection{Minimum machine volume as a complexity measure}
In the present article, we support the above philosophical solution to
the choice of the reference machine with basic observations. Let us
define physical message complexity:
\begin{equation}
  \label{eq:physmsg1}
  C_V(x) \triangleq \min\{  V(M) \ | \ M \rightarrow x \}
\end{equation}
where $x \in D^+$ is any d-ary message written in an alphabet $D$, $M$
is any physical machine (finite mechanism) that emits the message $x$
(denoted $M \rightarrow x$), and $V(M)$ is the volume of machine
$M$. $M$ is supposed to contain all physical computers that can emit
message $x$.

\prettyref{eq:physmsg1} is too abstract and it would have to be
connected to physical law to be useful. However, it allows us to
reason about the constraints we wish to put on physical complexity.
If we imagine what sort of device $M$ would be, $M$ is supposed to
contain every possible physical computer that can emit a message. For
this definition to be useful, the concept of emission would have to be
determined. Imagine for now that the device emits photons that can be
detected by a sensor, interpreting the presence of a photon with
frequency $f_i$ as $d_i \in D$. 
%One would first have to build a device
%that can output a single photon and stops before outputting any
%messages.
It might be hard for us to build the
\emph{minimal} device that can do this. However, let us assume that
such a device can exist and be simulated.  
It is likely that this minimal hardware 
would occupy quite a large volume compared to the output it emits.
With every added unit of message complexity, the minimal device would
have to get larger. We may consider additional complications. For
instance, we may demand that these machines do not receive any
physical input, i.e., supply their own energy, which we call a
\emph{self-contained} mechanism. We note that resource bounds can also
be naturally put into this picture.

When we use $C_V(x)$ instead of $H_U(x)$, we do not only eliminate the
need for a reference machine, but we also eliminate many constraints
and constants in AIT. First of all, there is not the same worry of a
self-delimiting program, because every physical machine that can be
constructed will either emit a message or not in isolation, although
its meaning slightly changes and will be considered in the
following. Secondly, we expect all the basic theorems of AIT to hold,
while the arbitrary constants that correspond to glue code to be
eliminated or minimized. Recall that the constants in AIT usually
correspond to such elementary operations as function composition and
so forth. Let us consider the sub-additivity of information which
represents a good example:
% \begin{equation}
%  \label{eq:subadd}
$H_U(x,y) = H_U(x) + H_U(y|x) + O(1)$
%\end{equation}
% The program that corresponds to the right-hand size, which generates
% bit string $xy$, may be considered to be \code{(defun ($p_1$) \dots)
% (defun ($p_2$) \dots) (BIT-CONCAT ($p_1$) ($p_2$ $p_1$))} for a
% LISP-based algorithmic complexity definition where $p_1$ generates
% $x$ and $p_2$ generates $y$ given an optimal program for $x$, and
% \code{BIT-CONCAT} would concatenate bit strings. The program schema
% illustrates the correctness of \prettyref{eq:subadd}; there is a
% small constant program that needs to be added to $p_1$ and $p_2$ in
% order to obtain a near-optimal code for $xy$.  Such a definition is
% natural in a LISP-like language and is among the best that we can
% have among choices of reference machines (note that Chaitin's
% definition of LISP-based complexity is different than this one).
When we consider $C_V(x,y$), however, the sub-additivity of information
becomes exactly
% \begin{equation}
%  \label{eq:phys-subadd}
$C_V(x,y) = C_V(x) + C_V(y|x)$
%\end{equation}
since there does not need to be a gap between a machine emitting a
photon and another sensing one. In the consideration of an underlying
physical theory of computing (like quantum computing), the relations
will further change, and become ever clearer.

% In the following, we discuss how the sub-additivity of algorithmic
% information changes with respect to assumed physical model of
% computation.

% We consider a popular model in digital physics which is a RUCA
% (Reversible Universal Cellular Automaton). Let $RUCA_1$ be a 1-D
% $d$-ary RUCA, with unspecified rules, but without gliders.  In such
% a crude model of digital physics, \prettyref{eq:phys-subadd} does
% not hold, instead we have a slightly different result.
% % \begin{lemma}
%   If $U=RUCA_1$, then
%   \begin{equation}
%   \label{eq:phys-subadd-2}
%   C(x,y) = C(x) + C(y|x) + O(1)
% \end{equation}
% % \end{lemma}

\subsection{Volume based algorithmic probability}

From the viewpoint of AI theory, however, what we are interested in is
whether the elimination of a reference machine may improve the
performance of machine learning. Recall that the convergence theorem
is related to the algorithmic entropy of the stochastic source with
respect to the reference machine. A reasonable concern in this case is
that the choice of a ``bad'' reference machine may inflate the errors
prohibitively for small data size, for which induction works best,
i.e., as the composition of a physical system may be poorly reflected
in an artificial language, increasing generalization error. On the
other hand, setting $U$ to the universe obtains an objective
measurement, which does not depend on subjective choices, and
furthermore, always corresponds well to the actual physical complexity
of the stochastic source. We shall first need to re-define algorithmic
probability for an alphabet of $D$. We propose using the exponential
distribution for a priori machine probabilities, which would be
applicable to any choice of real-valued units, although we would favor
Planck-units and integer measurements of volume.
\begin{equation}
  \label{eq:alp-phys}
  P(x) \triangleq  \frac{\sum_{M \rightarrow xD^*} \lambda e^{-\lambda V(M)}}
  {\sum_{M \rightarrow D^+} \lambda e^{-\lambda V(M)}}
\end{equation}
An unbiased choice for parameter $\lambda$ here would be $1$, however,
for physical reasons a smaller parameter may be preferable.  Here, it
does not matter that any machine-encodings of $M$ are prefix-free,
because infinity is not a valid concern in physical theory, and any
arrangement of quanta is possible (although not stable). Due to
general relativity, there cannot be any influence from beyond the
observable universe, i.e., there is not enough time for any message to
arrive from beyond it, even if there is anything beyond the cosmic
horizon. Therefore, the volume $V(M)$ of the largest machine is
constrained by the volume of the observable universe, i.e., it is
finite. Hence, the sums always converge.

\subsection{Minimum machine energy and action}

We now propose alternatives to minimum machine volume complexity.
While volume quantifies the initial space occuppied by a machine,
\emph{energy} accounts for every aspect of operation. In general relativity,
the energy distribution determines the curvature of space-time, and
energy is equivalent to mass via creation and annihilation of
particle-antiparticle pairs. Likewise, the unit of $h$ is $J.sec$,
i.e., energy-time product, quantum of \emph{action} and quantifies dynamical evolution of physical systems. 
Let $C_E(x) \triangleq
\min\{ E(M) \ | \ M \rightarrow x \}$ be the energy complexity of
message, and $C_A(x) \triangleq \min\{ A(M) \ | \ M \rightarrow x \}$
action (or action volume $E.t$) complexity of message which quantify the computation
and transmission of message $x$ by a finite mechanism \cite{margolus-levitin}. Further variants may be construed by considering how much energy and action it
takes to build $M$ from scratch, which include the work
required to make the constituent quanta, and are called constructive
energy $C_{Ec}(x)$ and action $C_{Ac}(x)$ complexity of messages,
respectively. Measures may also be defined to account for
machine construction, and message transmission, called total
energy $C_{Et}(x)$, and total action $C_{At}(x)$ complexity of messages. Versions
of algorithmic probability may be defined for each of these six new
complexity measures in similar manner to \prettyref{eq:alp-phys}. Note that the trick in algorithmic probability is maximum uncertainty about the source $\mu$. For energy based probability, if $\mu$ is at thermal equilibrium we may thus use the Boltzmann distribution $P(M) = e^{-E/kT}$ for a priori machine probabilities instead of the exponential distribution, which also maximizes uncertainty. We may also model a priori probabilities with a canonical ensemble, using $P(M) = e^{(F-E)/kT}$ where $F$ is the Helmholtz free energy.

\subsection{Restoring subjectivity}
% TODO: how to handle prior information
Solomonoff's observation that subjectivity is required to solve any
problem of significant complexity is of paramount importance. Our
proposal of using a physical measure of complexity for objective
inference does not neglect that property of universal
induction. Instead, we observe that a guiding pdf contains prior
information in the form of a pdf. Let $U_1$ be a universal computer
that contains much prior information about a problem domain, based on
a universal computer $U$ that does not contain any significant
information.  Such prior information may always be split off to a
memory bank.
\begin{equation}
  \label{eq:1}
  P_{U_1}(x) = P_U(x | M)
\end{equation}
Therefore, we can use a conditional physical message complexity given
a memory bank to account for prior information, instead of modifying a
pdf.  Subjectivity is thus retained. Note that the universal induction
view is compatible with a Bayesian interpretation of probability,
while admitting that the source is real, which is why we can eliminate
the bias about reference machine -- there is a theory of everything
that accurately quantifies physical processes in this universe.

Choosing the universe as $U$ has a particular disadvantage of using
the lowest possible level computer architecture. Science has not yet
formulated complete descriptions of the computation at the lowest
level of the universe, therefore further research is needed. However,
for solving problems at macro-scale, and/or from artificial sources,
algorithmic information pertaining to such domains must be encoded as
prior information in $M$, since otherwise solution would be
infeasible.

\subsection{Quantum algorithmic probability and physical models}

Note that it is well possible to extend the proposal in this section
to a quantum version of AIT by setting $U$ to a universal quantum
computer. There are likely other advantages of using a universal
quantum computer, e.g., efficient simulation of physical systems. For
instance, the quantum circuit model may be used, which seems to be
closer to actual quantum physical systems than Quantum
Turing Machine model \cite{miszczak:qc}. A universal quantum computer
model will also extend the definition of message to any quantum
measurement.  In particular, we the input to the quantum circuit is
$\ket{0\dots}$ (null) while the output is the quantum measurement of
message $\ket{x}$. Since quantum computers are probabilistic, multiple
trials must be conducted to obtain the result with high
probability. 
Also, Grover's algorithm may be applied to accelerate
universal induction approximation procedures.

% \subsection{Physical model classes}

% All that we have explained with regards to the choice of the
% reference machine must be understood in the sense that the models
% Our proposal must be understood in the sense that an induction
% system prefer physical models wherever possible.
All physical systems do reduce properly to quantum systems, however,
only problems at the quantum-scale would require accurate simulation
of quantum processes.  An ultimate AGI system would choose the
appropriate physical model class for the scale and domain of sensor
readings it processes. Such a machine would be able to adjust its
attention to the scale of collisions in LHC, or galaxy clusters
according to context. This would be an important ability for an
artificial scientist, as different physical forces are at play at
different scales; nature is not uniformly scale-free, although some
statistical properties may be invariant across scales. The formalism
of phase spaces and stochastic dynamical systems may be used to
describe a large number of physical systems. What matters is that a
chosen physical formalism quantifies basic physical resources in a way
that allows us to formulate physical complexity measures. We contend
however that a unified language of physics is possible, in accordance
with the main tenets of logical empiricism.
% We merely argued of objectivity that the stochastic models used
% should be physical models, and we can posit a priori probabilities
% based on physical properties, not arbitrary, abstract properties
% independent of physics.

\subsection{The physical semantics of halting probability}
The halting probability $\Omega_U$ is the probability that a random
program of $U$ will halt, and it is semi-computable much like
algorithmic probability. What happens when we set $U$ to the universe?
We observe that there is an irreducible mutual algorithmic information
between any two stochastic sources, which is the physical law, or the
finite set of axioms of physics (incomplete presently). This
irreducible information corresponds to $U$ in our framework, and it is
equivalent to the uniformity of physical law in cosmology for which
there is a wealth of evidence \cite{Tubbs:Uniformity}. It is known
that $\Omega_U$ contains information about difficult conjectures in
mathematics as most can be transformed to instances of the halting
problem. Setting $U$ to a (sufficiently complete) theory of physics
biases $\Omega_U$ to encode the solutions of non-trivial physical
problems in shorter prefixes of its binary expansion, while it still
contains information about any other universal machines and problems
stated within them, e.g., imaginary worlds with alternative physics.

\section{Discussion}

\subsection{Dissolving the problem of induction}

The problem of induction is an old philosophical riddle that we cannot
justify induction by itself, since that would be circular. 
% Indeed,
% this is a common confusion among those philosophers who are not
% familiar with the theory of universal computer.
 If we follow the
proposed physical message complexity idea, for the first capable
induction systems (brains) to evolve, they did not need to have an a
priori, deductive proof of induction. However, the evolution process
itself works inductively as it proceeds from simpler to more complex
forms which constitute and expend more physical entropy. Therefore,
induction does explain how inductive systems can evolve, an
explanation that we might call a glorious recursion, instead of a
vicious circle: an inductive system can invent an induction system
more powerful than itself, and it can also invent a computational
theory of how itself works when no such scientific theory previously
existed, which is what happened in Solomonoff's brain.

\subsection{Disproving Boltzmann brains}
The argument from practical finiteness of the universe was mentioned
briefly by Solomonoff in \cite{sol67}.
% the following passage
% \begin{quote}
%   The questions naturally arise, ``are finitely describable
%   stochastic sequences of any importance?'' ``Do they often occur in
%   the world about us?''
%
%   In answer to these questions, it is difficult to conceive of any
%   sequence of data that is not of this form. Suppose that the laws
%   of the universe are finite in number, and that there are a finite
%   number of particles in the universe. The laws can be statistical
%   and/or deterministic.
%
%   If one describes in a finite number of words, where one has
%   obtained a data sequence from such a universe, then that data
%   itself will be a finitely describable stochastic sequence.
%
%   If there are an infinite number of particles in the universe, I
%   think the previous statement is still true, but I am not certain.
%
%   Most, if not all, of the theorizing in physics has tacitly assumed
%   the finite describability of the laws of the universe.
% \end{quote}
Let us note, however, that the abstract theory of algorithmic
probability implies an infinite probabilistic universe, in which every
program may be generated, and each bit of each program is
equiprobable. In such an abstract universe, a Boltzmann Brain, with
considerably more entropy than our humble universe is even possible,
although it has a vanishingly small probability. In a finite observable
universe with finite resources, however, we obtain a slightly
different picture, for instance any Boltzmann Brain is improbable, and a
Boltzmann Brain with a much greater entropy than our universe would be
impossible ($0$ probability).  Obviously, in a sequence of universes with increasing volume of observable universe, the limit would be much like pure
algorithmic probability.  However, for our definition of physical
message complexity, a proper physical framework is much more
appropriate, and such considerations quickly veer into the territory
of metaphysics (since they truly consider universes with physical law
unlike our own). Thus firmly footed in contemporary physics, we gain a
better understanding of the limits of ultimate intelligence.

\subsection{Refuting the Platonist objection to algorithmic information}
An additional nice property of using physical stochastic models, e.g.,
statistical mechanics, stochastic dynamical systems, quantum computing
models, instead of abstract machine or computation models is that we
can refute a well-known objection to algorithmic information by
Raatikainen \cite{raatikainen}, which depends on unnatural
enumerations of recursive functions, essentially constructing
reference machines with a lot of useless information. Such superfluous
reference machines would incur a physical cost in physical message
complexity, and therefore they would not be picked by our definition,
which is exactly why you cannot shuffle program indices as you like,
because such permutations require additional information to encode. An
infinite random shuffling of the indices would require infinite
information, and impossible in the observable universe, and any
substantial reordering would incur inordinate physical cost in a
physical implementation of the reference
machine. %Raatikainen insists in his paper that ``one can say that acceptable systems %of indices provide the same structure theory for recursive functions as the %standard one. Thus, from the point of view of computatiblity, it does not %really make any difference which acceptable system of indices one uses.'',
% w hich we just showed to be a naive misunderstanding of theory of
% computation, which is essentially a physical theory, not a theory
% about %massless, energy-free, functions in a Platonic realm.
Raatikainen contends that his self-admittedly bizarre and unnatural
constructions are fair play because a particular way of representing
the class of computable functions cannot be privileged.
% Of course, that contradicts with a proper scientific world-view.
Better models of computation accurately measure time, space and energy
complexities of physical devices, which is why they \emph{are}
privileged.  RAM machine model is a better model of personal computers
with von Neumann architecture than a Turing Machine, which is preferable to 
a model with no physical complexity measures.
%Still, a Turing
%Machine is empirically much better than a model that
%does not quantify physical resources.
% In other words, Raatikainen's misunderstanding owes to his
% misunderstanding of the world at large; he apparently believes that
% mathematical objects
% exist %in a Platonic realm, and by extension, computations, which are also %mathematical, are not essentially physical.

\subsection{Concluding remarks and future work}

We have introduced the basic philosophical problems of an
investigation into the ultimate limits of intelligence.  We have
covered a very wide philosophical terrain of physical considerations
of completeness and objective choice of reference machine, and we have
proposed several new kinds of physical message complexity and
probability.
% We have interpreted halting probability, the problem of induction,
% Boltzmann brains, and Platonist objections in the context of
% physical, %objective reference machines.
Much work remains to fully connect existing body of physical theory to
algorithmic probability.
% We anticipate that there might be interesting bridge theorems to be
% obtained.  Detailed investigation of employing universal models of
% quantum computation shall be pursued in successive publications of
% the ultimate intelligence research program.

% \bibliographystyle{plain}
\tiny{ \bibliographystyle{splncs03}
  \bibliography{agi,physics,complexity} }

\end{document}